\documentclass[10pt,twocolumn,letterpaper]{article}

\usepackage{cvpr}
\usepackage{times}
\usepackage{epsfig}
\usepackage{subcaption}
\usepackage{graphicx}
\usepackage{amsmath}
\usepackage{amssymb}
\usepackage{soul}

\def\*#1{\mathbf{#1}}

\newcommand{\mb}[1]{\mathbf{#1}}

\newcommand{\boldhead}[1]{\vspace{0.05in}\noindent\textbf{#1.}}

\usepackage[pagebackref=true,breaklinks=true,letterpaper=true,colorlinks,bookmarks=false]{hyperref}

\cvprfinalcopy 

\def\cvprPaperID{1625} 

\ifcvprfinal\fi
\begin{document}

\title{Predicting Complete 3D Models of Indoor Scenes}

\author{Ruiqi Guo \\
UIUC, Google
\and Chuhang Zou \\
UIUC
\and Derek Hoiem \\
UIUC
}

\maketitle

\begin{abstract}
One major goal of vision is to infer physical models of objects, surfaces, and their layout from sensors.  In this paper, we aim to interpret indoor scenes from one RGBD image.  Our representation encodes the layout of walls, which must conform to a Manhattan structure but is otherwise flexible, and the layout and extent of objects, modeled with CAD-like 3D shapes.  We represent both the visible and occluded portions of the scene, producing a $complete$ 3D parse.  Such a scene interpretation is useful for robotics and visual reasoning, but difficult to produce due to the well-known challenge of segmentation, the high degree of occlusion, and the diversity of objects in indoor scene.  We take a data-driven approach, generating sets of potential object regions, matching to regions in training images, and transferring and aligning associated 3D models while encouraging fit to observations and overall consistency.  We demonstrate encouraging results on the NYU v2 dataset and highlight a variety of interesting directions for future work.
\end{abstract}

\section{Introduction}
\label{sec:intro}

Recovering the layout and shape of surfaces and objects is a foundational problem in computer vision.
Early approaches, such as reconstruction from line drawings~\cite{roberts1963machine}, attempt to infer 3D object and surface models based on shading cues or boundary reasoning. But the complexity of natural scenes is too difficult to model with hand-coded processing and rules.  More recent approaches to 3D reconstruction focus on producing detailed literal geometric models, such as 3D point clouds or meshes, from multiple images~\cite{furukawa2009iccv}, or coarse interpreted models, such as boxy objects within a box-shaped room~\cite{hedau2009iccv}, from one image.

\begin{figure}[!ht]
\centering
\includegraphics[width=\columnwidth]{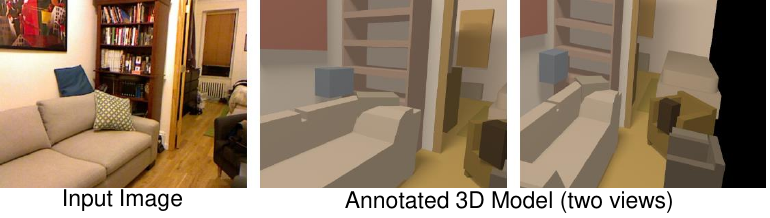}
\vspace{-0.25in}
\caption{
\label{fig:intro}
Our goal is to recover a 3D model (right) from a single RGBD image (left), consisting of the position, orientation, and extent of layout surfaces and objects.}
\vspace{-0.2in}
\end{figure}

\begin{figure*}[!ht]
\centering
\includegraphics[width=0.75 \textwidth]{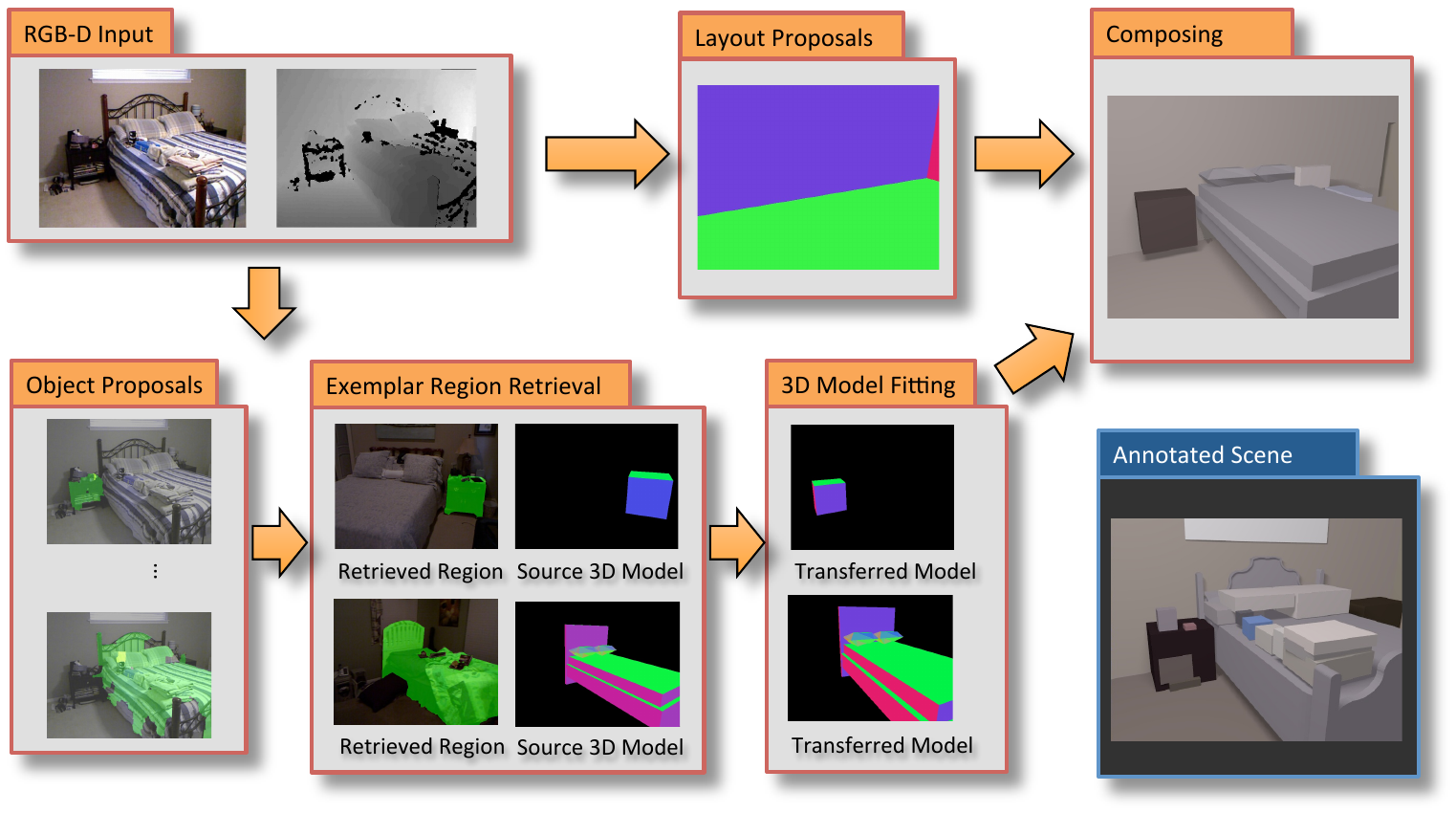}
\vspace{-0.15in}
\caption{
\label{fig:framework}
\textbf{Framework. } Given an input RGB-D (upper-left), we propose possible layouts and object regions. Each object region is matched to regions in the training set, and corresponding 3D model exemplars are transferred and aligned to the input depth image. The subset of proposed objects and walls is then selected based on consistency with observed depth, coverage, and constraints on occupied space. We show an example result (upper-right) and ground truth annotations (lower-right).
}
\vspace{-0.1in}
\end{figure*}

This paper introduces an approach to recover complete 3D models of indoor objects and layout surfaces from an RGBD (RGB+Depth) image (Fig.~\ref{fig:intro}), aiming to bridge the gap between detailed literal and coarse interpretive 3D models. Recovering 3D models from images is highly challenging due to three ambiguities: the loss of depth information when points are projected onto an image; the loss of full 3D geometry due to occlusion; and the unknown separability of objects and surfaces.  In this paper, we choose to work with RGBD images, rather than RGB images, so that we can focus on designing and inferring a useful representation, without immediately struggling with the added difficulty of interpreting geometry of visible surfaces.  Even so, ambiguities due to occlusion and unknown separability of nearby objects make 3D reconstruction impossible in the abstract.  If we see a book on a table, we observe only part of the book's and table's surfaces.  How do we know the full shape of them, or even that the book is not just a bump on the table?  Or how do we know that a sofa does not occlude a hole in the wall or floor?  We don't.  But we expect rooms to be enclosed, typically by orthogonal walls and horizontal surfaces, and we can guess the extent of the objects based on experience with similar objects.  Our goal is to provide computers with this same interpretive ability.

\boldhead{Scene Representation}
We want an expressive representation that supports robotics (e.g., where is it, what is it, how to move around and interact), graphics (e.g., what would the scene look like with or without this object), and interpretation (e.g., what is the person trying to do).  Importantly, we want to show what is in the scene and what could be done, rather than only a labeling of visible surfaces.  In this paper, we aim to infer a 3D geometric model that encodes the position and extent of layout surfaces, such as walls and floor, and objects such as tables, chairs, mugs, and televisions.  In the long term, we hope to augment this geometric model with relations (e.g., this table supports that mug) and attributes (e.g., this is a cup that can serve as a container for small objects and can be grasped this way).
Adopting the annotated representation from Guo and Hoiem~\cite{guo2013iccv}, we represent layout surfaces with 3D planar regions, furniture with CAD exemplars, and other objects with coarser polygonal shapes.  Our goal is to automatically determine the position, orientation, and extent of layouts and objects.  We measure success primarily according to accuracy of depth prediction of complete layout surfaces and occupancy (voxel) accuracy of the entire model, but also include evaluations of labeling and segmentation accuracy for ease of comparison by any future work.

\boldhead{Contributions}
Our primary contribution is an approach to recover a 3D model of room layout and objects from an RGBD image.  A major challenge is how to cope with the huge diversity of layouts and objects.
Rather than restricting to a parametric model and a few detectable objects, as in previous single-view reconstruction work, our models encode every layout surface and object with 3D models that approximate the original depth image under projection.  The flexibility of our models is enabled through our approach (Fig.~\ref{fig:framework}) to propose a large number of likely layout surfaces and objects and then compose a complete scene out of a subset of those proposals while accounting for occlusion, image appearance, depth, and layout consistency.
In contrast to multiview 3D reconstruction methods, our approach recovers complete models from a limited viewpoint, attempting to use priors and recognition to infer occluded geometry, and parses the scene into individual objects and surfaces, instead of points, voxels, or contiguous meshes.  Thus, in some sense, we provide a bridge between the goals of interpretation from single-view and quantitative accuracy from multiview methods.  That said, many challenges remain.



\section{Related Work}

\textbf{Room layout} is often modeled as a 3D box (cuboid)~\cite{hedau2009iccv,flint2011iccv,schwing2012eccv,zhang2013iccv,zhang2014eccv}.  A box provides a good approximation to many rooms, and it has few parameters so that accurate estimation is possible from single RGB images~\cite{hedau2009iccv,flint2011iccv,schwing2012eccv} or panoramas~\cite{zhang2014eccv}.  Even when a depth image is available, a box layout is often used (e.g.,\cite{zhang2013iccv}) due to the difficulty of parameterizing and fitting more complex models.  
Others, such as~\cite{delage2006cvpr,lee2009cvpr}, estimate a more detailed Manhattan-structured layout of perpendicular walls based on visible floor-wall-ceiling boundaries.
Methods also exist to recover axis-aligned, piecewise-planar models of interiors from large collections of images~\cite{furukawa2009iccv} or laser scans~\cite{xiao2012eccv}, benefitting from more complete scene information with fewer occlusions.  We model the walls, floor, and ceiling of a room with a collection of axis-aligned planes with cutouts for windows, doors, and gaps, as shown in Figure~\ref{fig:layoutProc}.  Thus, we achieve similar model complexity to other methods that require more complete 3D measurements.

\textbf{3D objects} are also often modeled as 3D boxes when estimating from RGB images~\cite{hedau2010eccv,xiao12nips,zhao2013cvpr,lin2013iccv,zhang2014eccv} or RGB-D images~\cite{lin2013iccv}.  But cuboids do not provide good shape approximations to chairs, tables, sofas, and many other common objects.  Another approach is to fit CAD-like models to depicted objects. Within RGB images, Lim et al.~\cite{lim2013iccv,lim2014eccv} find furniture instances, and Aubry et al.~\cite{aubry2014cvpr} recognize chairs using HOG-based part detectors. In RGB-D images, Song and Xiao~\cite{song2014eccv} search for chairs, beds, toilets, sofas, and tables by sliding 3D windows and enumerating all possible poses.
Our approach does not aim to categorize objects but to find an approximate shape for any object, which could be from a rare or nonexistent category in the training set.
We take an exemplar-based approach, conjecturing that a similar looking object from a different category can often still provide a good approximation to 3D extent.  Incorporating category-based 3D detection, such as~\cite{song2014eccv}, is a promising direction for future work.

Our use of \textbf{region transfer}
is inspired by the SuperParsing method of Tighe and Lazebnik~\cite{tighe2010eccv}, which transfers pixel labels from training images based on retrieval.  Similar ideas have also been used in other modalities: Karsch et al.~\cite{karsch2012eccv} transfers depth, Guo and Hoiem~\cite{guo2012eccv} transfers polygons of background regions, and Yamaguchi et al.~\cite{yamaguchi2013iccv} transfers clothing items.  Exemplar-based 3D modeling is also employed by Satkin and Hebert~\cite{satkin2013iccv} to transfer 3D geometry and object labels from entire scenes.  Rather than retrieving based on entire images (which is too constraining) or superpixels (which may not correspond to entire objects), we take an approach of proposing a bag of object-like regions and resolving conflicts in a final compositing process that accounts for fidelity to observed depth points, coverage, and consistency.  In that way, our approach also relates to work on segmentation~\cite{gupta2013cvpr,dollar2013iccv} and parsing~\cite{ren2012cvpr,banica2013iccv} of RGBD images and generation of bags of object-like regions~\cite{carreira2012pami,endres2010eccv,manen2013iccv}.  We also incorporate a per-object 3D alignment procedure that reduces the need to match to objects with exactly the same scale and orientation.

\textbf{Other techniques} that have been developed for scene interpretation from RGB-D images do not provide 3D scene interpretations but could be used to improve or extend our approach.  For example, Silberman et al.~\cite{silberman2012eccv} infer support labels for regions, and Gupta et al.~\cite{gupta2014eccv} segment images into objects and assign category labels.  

\section{Approach}

We aim to find a set of layout and object models that fit RGB and depth observations and provide a likely explanation for the unobserved portion of the scene.  We can write this as
\begin{align*}
\{\mb M, \mb \theta\}=\arg \min (\rm{AppearanceCost}(\mathcal I_{RGBD},\mb M, \mb \theta)\\
+\rm{DepthCost}(\mathcal I_D, \mb M, \mb \theta)+\rm{ModelCost}(\mb M, \mb \theta)).
\label{eq:global}
\end{align*}
$\mathcal I_{RGBD}$ is the RGB-D image, $\mathcal I_D$ the depth image alone, $\mb M$ a set of 3D layout surfaces and object models, and $\mb \theta$ the set of parameters for each surface/object model, including translation, rotation, scaling, and whether the model is included.  $\rm{AppearanceCost}$ encodes that object models should match underlying region appearance, rendered objects should cover pixels that look like objects (versus layout surfaces), and different objects should have evidence from different pixels.  $\rm{DepthCost}$ encourages similarity between the rendered scene and observed depth image.  $\rm{ModelCost}$ penalizes intersection of 3D object models.  

We propose to tackle this complex optimization problem in stages (Figure~\ref{fig:framework}): (1) propose candidate layout surfaces and objects; (2) improve the fit of each surface/object model to the depth image; (3) choose a subset of models that best explains the scene.  Layout elements (wall, floor, and ceiling surfaces) are proposed by scanning for planes that match observed depth points and pixel labels and then finding boundaries and holes (Sec.~\ref{subsec:layout}).
The estimation of object extent is particularly difficult.  We propose an exemplar-based approach, matching regions in the input RGBD image to regions in the training set and transferring and aligning corresponding 3D models (Sec.~\ref{subsec:object}).  We then choose a subset of objects and layout surfaces that minimize the depth, appearance, and model costs using a specialized heuristic search (Sec.~ \ref{subsec:composition}).

\subsection{Dataset and Pre-processing}

We perform experiments on the NYU v2 dataset collected by Silberman et al.~\cite{silberman2012eccv}, using the standard training/test split.  The dataset consists of 1449 RGB-D images, with each image segmented and labeled into object instances and categories.  Each segmented object also has a corresponding annotated 3D model, provided by Guo and Hoiem~\cite{guo2013iccv}.  As detailed in Sec.~\ref{subsec:object}, we use object category labels as proxy for similarity when training the region matching function and transfer the annotated 3D object models, which consist of Sketchup exemplars for furniture and extruded polygons for other objects.  Given a test image, we use the code from~\cite{silberman2012eccv} to obtain an oversegmentation with boundary likelihoods and the probability that each pixel $j$ corresponds to an object ${\rm P_{object}}(j;\mathcal I_{RGBD})$ (versus wall, floor, or ceiling).  We use the code from~\cite{guo2013iccv} to find the major orthogonal scene orientations, used to align the scene and search for layout planes.

%
%

\subsection{Layout Proposals}
\label{subsec:layout}

\begin{figure}[t]
\begin{center}
\includegraphics[width=0.7\columnwidth]{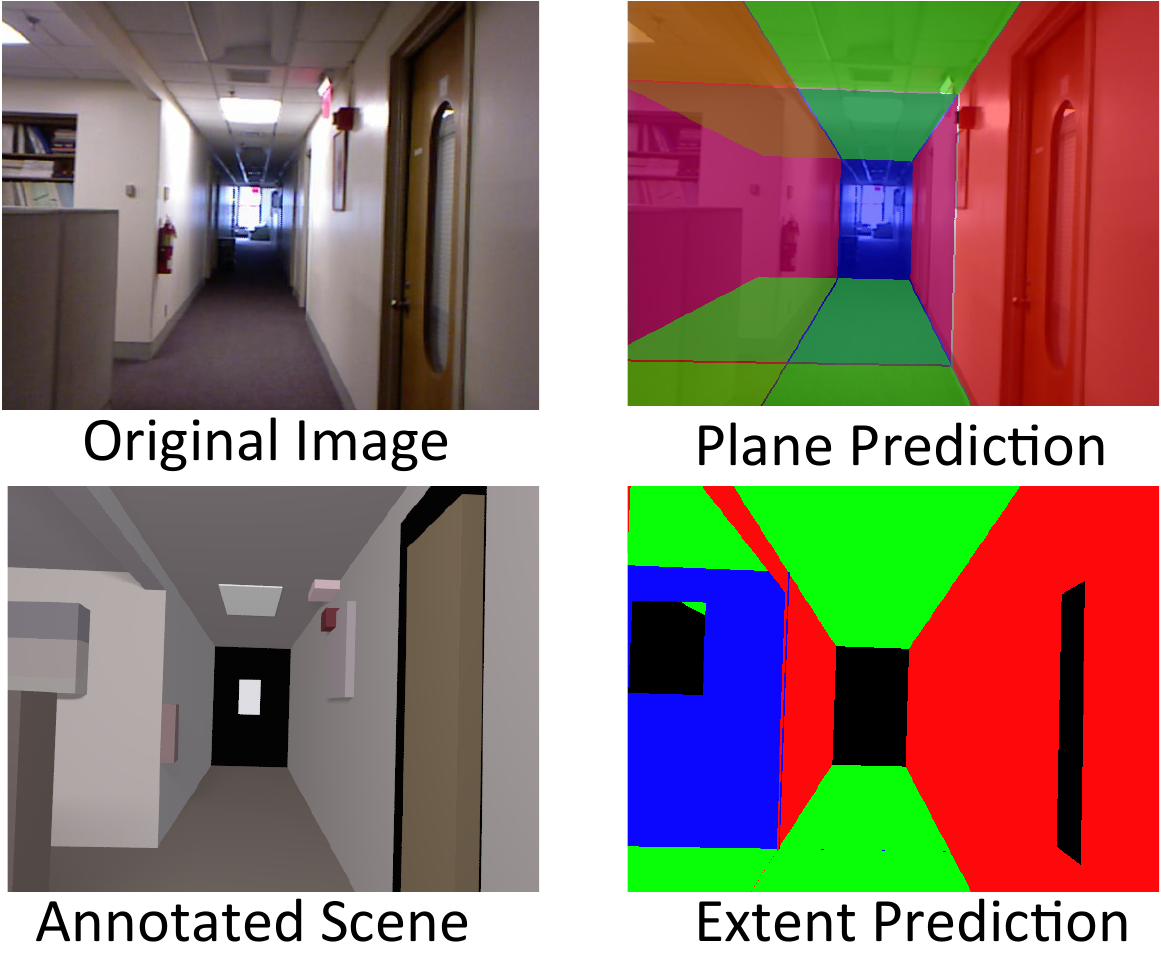}
\end{center}
\vspace{-0.25in}
\caption{\label{fig:layoutProc}
\textbf{Layout proposal}. We detect axis-aligned horizontal and vertical planes in the input depth image and estimate the extent of each surface.  Human-annotated layout is on the lower-left.}
\vspace{-0.1in}
\end{figure}

We want to estimate the full extent of layout surfaces, such as walls, floor, and ceiling.  The layouts of these surfaces can be complex.  For example, the scene in Figure~\ref{fig:layoutProc} has several ceiling-to-floor walls, one with a cutout for shelving, and a thin strip of wall below the ceiling on the left.  The cabinet on the left could easily be mistaken for a wall surface.  Our approach is to propose a set of planes in the dominant room directions.  These planes are labeled into ``floor'', ``ceiling'', ``left wall'', ``right wall'', or ``front wall'' based on their position and orientation. Then, the extent of the surface is determined based on observed depth points.

To find layout planes, we aggregate appearance, depth, and location features and train a separate linear SVM classifier for each of the five layout categories to detect planes.  We define ${\rm p}(p_i ; P)=\mathcal N(dist(p_i, P), \sigma_p) \mathcal N(dist(n_i,P), \sigma_n)$ as the probability that a point with position $p_i$ and normal $n_i$ belongs to plane $P$.  $\mathcal N$ is a zero-mean normal distribution, $dist(p_i,P)$ is the Euclidean point to plane distance divided by distance to the point, and $dist(n_i,P)$ is the angular surface normal difference.  We set $\sigma_p=0.025$ and $\sigma_n=0.0799$ are based on Kinect measurement error. Each pixel also has a probability of belonging to floor, wall, ceiling, or object, using code from~\cite{silberman2012eccv}.  The plane detection features are $f_1=\sum_i{{\rm p}(p_i ; P)}$; $f_2...f_5$, the sum in $f_1$ weighted by each of the four label probabilities; $f_6$, the number of points behind the plane by at least 3\% of plane depth; $f_7...f_{11}=(f_1...f_5)/f_6$; and $f_{12}$, a plane position prior estimated from training data.  Non-maximum suppression is used to remove weaker detections within 0.15m of a stronger detected plane.  Remaining planes are kept as proposals if their classification score is above a threshold, typically resulting in 4-8 layout surfaces.

The maximum extent of a proposed plane is determined by its intersection with other planes: for example, the floor cuts off wall planes at the base.  Further cut-outs are made by finding connected components of pixels with depth 5\% behind the plane, projecting those points onto the plane, fitting a bounding box to them, and removing the bounding box from the plane surface.  Intuitively, observed points behind the plane are evidence of an opening, and the use of a bounding box enforces a more regular surface that hypotheses openings behind partly occluded areas.


\subsection{Object Proposals}
\label{subsec:object}
\subsubsection{Region Proposals}

\begin{figure}
\includegraphics[width=\columnwidth]{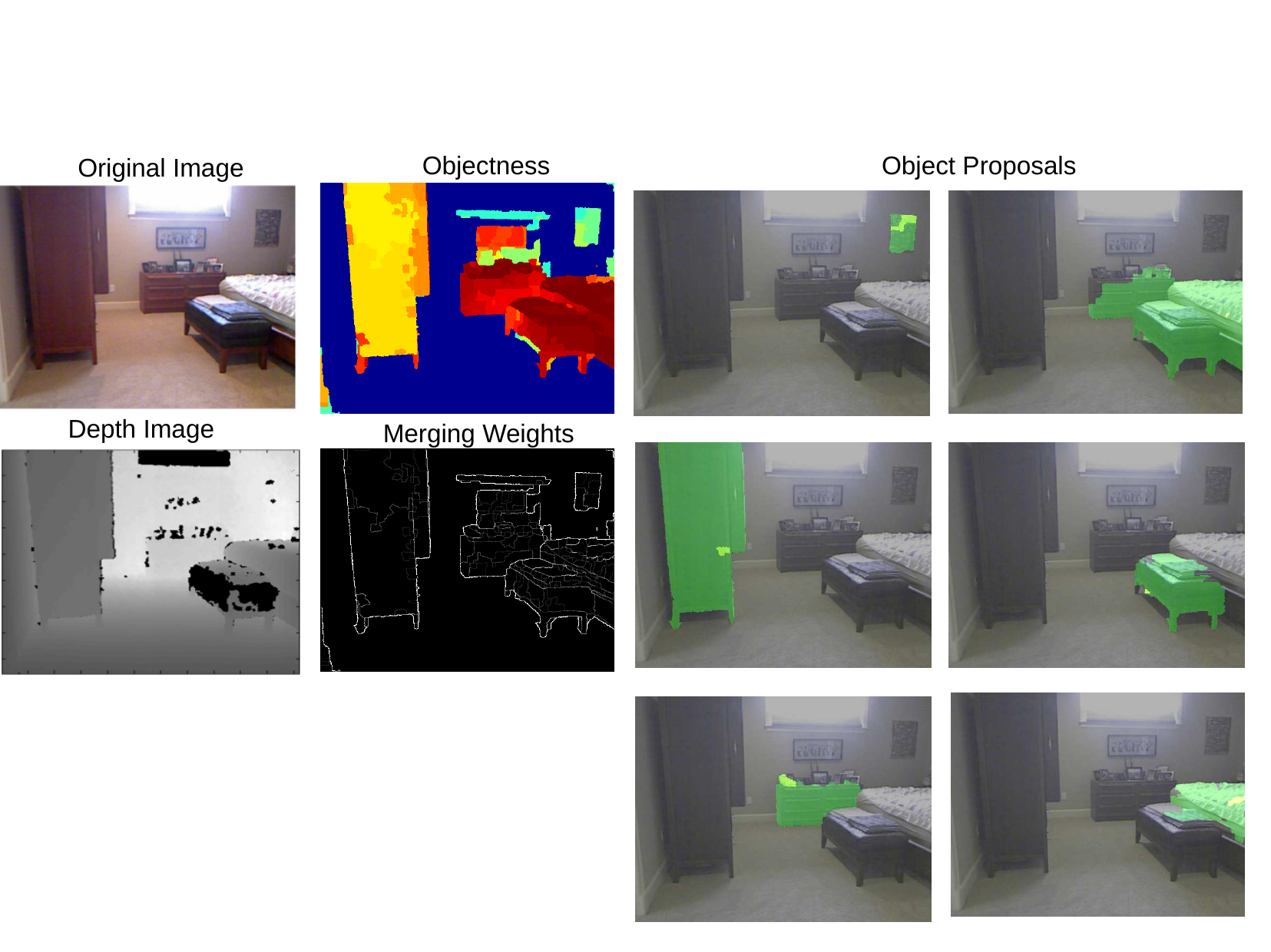}
\vspace{-0.2in}
\caption{
\label{fig:proposal}
\textbf{Object proposals.}
On the left is the original RGB-D image. The 2nd column from top: boundary strength; the probability of foreground object from parsing~\cite{silberman2012eccv}; the merging costs weighted by objectness. The 3rd and 4th columns show object proposals from our approach (light green=seed region; green=full proposed region).}
\vspace{-0.1in}
\end{figure}

Often, objects within a room are tightly clustered in 3D, so that parsing is difficult, even with a depth map.  Therefore, our approach, visualized in Figure~\ref{fig:proposal}, is to propose regions that may correspond to objects.  We start with an oversegmentation and boundary strengths provided by the code of Silberman et al.~\cite{silberman2012eccv}. We create a neighborhood graph, with superpixels as nodes and boundary strength as weight connecting adjacent superpixels.  We then apply the randomized Prim's algorithm~\cite{manen2013iccv} on this graph to obtain a set of region proposals. The seed region of the Prim's algorithm is sampled according to the objectness of the segment, so that the segments that are confidently layout are never sampled as seeds. We also sample for size constraints and merging threshold to produce a more diverse set of segmentations. We suppress regions that are near-duplicates of other regions to make the set of proposals more compact.  In our experiments, we generate 100 candidate object regions.

Other proposal mechanisms, such as~\cite{carreira2012pami,endres2010eccv,gupta2014eccv,krahenbul2014eccv}, could also be used and some likely would be superior to the proposals based on Silberman et al., which we use for simplicity of code base and exposition. Our idea of using object likelihood to focus the proposals would also likely be useful for these other proposal methods.

\subsubsection{3D Region Retrieval}

We use candidate object regions to retrieve similar objects and transfer the associated 3D models. 
Given a region from the segmentation, we extract both the 2D appearance and 3D shape features, and then retrieve similar regions from the training set using nearest-neighbor search, in the same spirit as super-parsing~\cite{tighe2010eccv} or paper doll parsing~\cite{yamaguchi2013iccv}. The difference is that (1) we transfer 3D object models, rather than semantic labels; (2) we make use of larger, possibly overlapping object proposals, versus smaller, disjoint superpixels; and (3) we learn a distance function to improve the retrieval, as explained below.  The features used for matching are described in Sec.~\ref{subsec:evalretrieval}.

Distance metric learning has been used to improve image nearest neighbor search~\cite{frome2007iccv}, and here we apply these approaches to object proposals. In the case of Mahalanobis distance, these approaches aim to learn a distance metric $\rm{dist}(\*x_i, \*x_j)$ between two feature vectors $\*x_i$ and $\*x_j$, defined as:
{\small
\[
\rm{dist}_{W}(\*x_i, \*x_j)=\sqrt{(\*x_i-\*x_j)^T\*W(\*x_i-\*x_j)}
\] }
and the metric is parameterized by a covariance matrix $\*W$. We use CCA which is often used to find embeddings of visual features to improve retrieval~\cite{gong2013ijcv}. CCA finds pairs of linear projections of the two views $\*\alpha^T\*X$ and $\*\beta^T\*Z$ that are maximally correlated. In our case, $\*X=\{\*x_i\}$ are the feature vectors of the regions and $\*Z=\{\*z_i\}$ is the label matrix, which contains the indicator vectors of the object category labels of the corresponding region.  Thus, we use object category as a proxy for object similarity.  The similarity weight matrix is then computed as $\*W=\*\alpha\*\alpha^T$.
We also tried triplet-based similarity learning from Frome et al.~\cite{frome2007iccv} and Schultz et al.~\cite{schultz2004nips}, but found that CCA performs better.  In our experiments, the ten object models nearest to each region proposal according to ${\rm dist_W}$ are retrieved.

\subsubsection{Fitting 3D object models}

Next, we need to align the retrieved 3D object models to fit the depth points of the proposal region in the target scene.  First, for each retrieved object, we initialize its location and scale based on the 3D bounding box of the source object and the region mask for the proposed object and then render the transferred object in 3D space to obtain the inferred 3D points.
The proposed model is often not the correct scale or orientation; e.g., a region of a left-facing chair often resembles a right-facing chair and needs to be rotated. We found that using Iterative Closest Point to solve for all parameters did not yield good results.
Instead, we enumerate six scales and five rotations from -90 to 90 degrees, perform ICP to solve for translation for each scale/rotation, and find the best transformation based on the following cost function:
{\small
\begin{align}
\label{eq:deptherr}
&{\rm FittingCost}(M_i, T_i)=\sum_{j\in r_i \cap s(M_i,T_i)} |\mathcal I_d(j)- \hat d(j;M_i,T_i)|\notag \\
&+\sum_{j\in r_i \cap \neg s(M_i,T_i)} C_{missing}\notag \\
&+\sum_{j\in \neg r_i \cap s(M_i,T_i)} \max(\hat d(j;M_i,T_i)-\mathcal I_d(j), 0)
\end{align}
}
where  $M_i$ is the model, $T_i$ the scale, rotation, and translation, $r_i$ the region proposal from the target scene, $s(.)$ the mask of the rendered aligned object, $\mathcal I_d(j)$ the observed depth at pixel $j$, and $\hat d(j)$ the rendered depth at $j$. The first term encourages depth similarity in the proposed region; the second penalizes pixels in the proposed region that are not rendered ($C_{missing}=0.3$ in our experiments); and the third term penalizes pixels in the rendered model that are closer than the observed depth (so the object does not stick out into space known to be empty).  For efficiency, we evaluate the ten retrieved object models before alignment based on this cost function, discard seven, and align the remaining three by choosing $T_i$ to minimize ${\rm FittingCost}(M_i, T_i)$.


\subsection{Scene Composition}
\label{subsec:composition}

Now, we have a set of candidate object and layout proposals and need to choose a subset that closely reproduce the original depth image when rendered, adhere to pixel predictions of object occupancy, and correspond to minimally overlapping 2D regions and 3D aligned models.  With the set of proposed models $\mb M$ and transformations $\mb T$ fixed, we need to solve for $\mb y$ with $y_i\in\{0,1\}$ indicating whether each layout or object model is part of the scene, minimizing Eq.~\ref{eq:global}:
\small
\begin{align}
\label{eq:global}
&{\rm selectionCost}(\mb y ; \mb M, \mb T)= \notag\\
&\sum_j {\rm clip}(\log_2(|\hat d(j;\mb M, \mb T,\mb y) - \mathcal I_d(j)|)-\log_2(1.03), [0~~1]) ~~+ \notag \\
&\sum_j |{\rm isObject}(j;\mb M,\mb T,\mb y)- {\rm P_{object}}(j; \mathcal I_{RGBD})| ~~+  \notag \\
&\sum_j \max(\sum_{i} y_i r_i(j)-1,0) ~~+ \notag \\
&\sum_{i,k>i} y_{i} y_{k} {\rm overlap3d}(M_{i},T_{i},M_{k},T_{k}).
\end{align}
\normalsize
The first term minimizes error between rendered model and observed depth.  We use log space so that error matters more for close objects.  We subtract $\log_2(1.03)$ and clip at 0 to 1 because errors less than 3\% of depth are within noise range, and we want to improve only reduction of depth if the predicted and observed depth is within a factor of 2.
The second term encourages rendered object and layout pixels to match the probability of object map from~\cite{silberman2012eccv}.  The third term penalizes choosing object models that correspond to overlapping region proposals (each selected object should have evidence from different pixels).  The second and third term, combined with region retrieval, account for the appearance cost between the scene model and observations.  The fourth term penalizes 3D overlap of pairs of aligned models, to encourage scene consistency.  For efficiency, we compute overlap based on the minimum and maximum depth of each object at each pixel.  $\hat d$ renders object models selected by $\mb y$, and $\rm isObject(j)=1$ identifies that a rendered pixel corresponds to an object, rather than a layout model.  Note that these terms each involve a cost per pixel in the 0 to 1 range, so they are comparable, and we found uniform weighting to work at least as well as others experimentally based on a grid search.

\begin{figure*}[!ht]
\centering
\includegraphics[width=\textwidth]{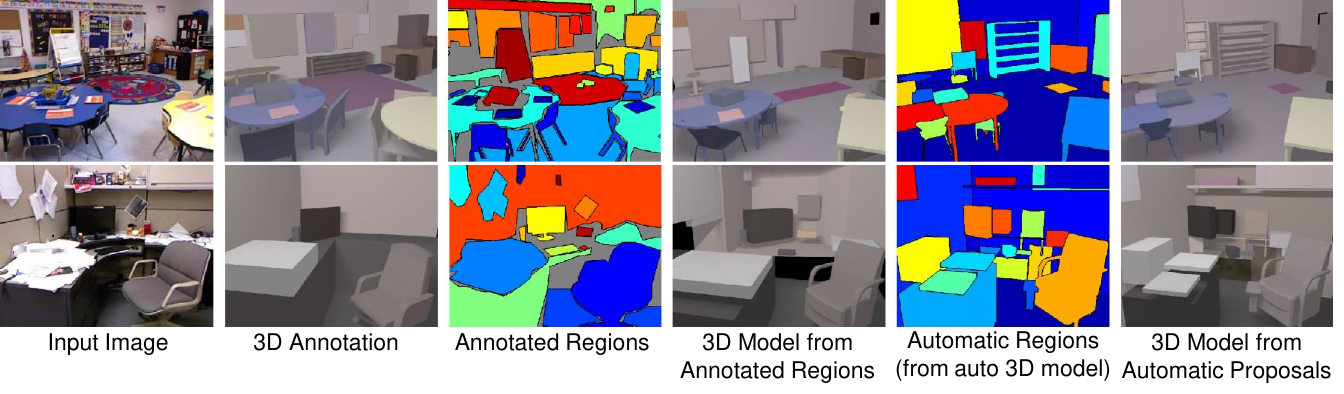}
\vspace{-0.25in}
\caption{
\label{fig:qual_long}
Examples of 3D models estimated based on manual segmentations or automatic segmentations.  Quality of 3D models from automatic region proposals is similar, due to effectiveness of generating multiple proposals and selecting those that best fit the scene.
}
\end{figure*}

The optimization is hard, primarily because the rendering in the first term depends jointly on $\mb y$.  For example, in Fig.~\ref{fig:qual_long} bottom, a sofa is heavily occluded by a table. Adding a single object that approximates both the table and sofa may reduce the cost by more than adding a sofa or table model by itself, but adding both the table and sofa models would give the best solution. We experimented linear programming relaxations, branch and bound, greedy search optimization, and approximations with pairwise terms, eventually determining the following algorithm to be most effective.  First, we set $\mb y= \mb 0$ and initialize with a greedy search on $\mb y$ to minimize the depth error (weighting the first term in Eq.~\ref{eq:global}) by a factor of 10).  We then perform a greedy search on all unweighted terms in Eq.~\ref{eq:global}, iteratively adding ($y_i=1$) or removing ($y_i=0$) the model that most reduces cost until no change yields further improvement.  Finally, for all layout proposals and a subset of object proposals that are not yet selected, try adding the proposed model, remove all models whose renderings overlap, and perform greedy search, keeping any changes that reduce cost.  The subset is the set of object models that are based on regions that were already selected or that were the second best choice for addition at any step of the greedy search.

In some experiments, we also use ground truth segmentations to examine impact of errors in region proposals.  In this case, we know that there should be one selected model per object region.  We find that a greedy search under this constraint provides good results.

\begin{figure*}[!ht]
\centering
\includegraphics[width=\textwidth]{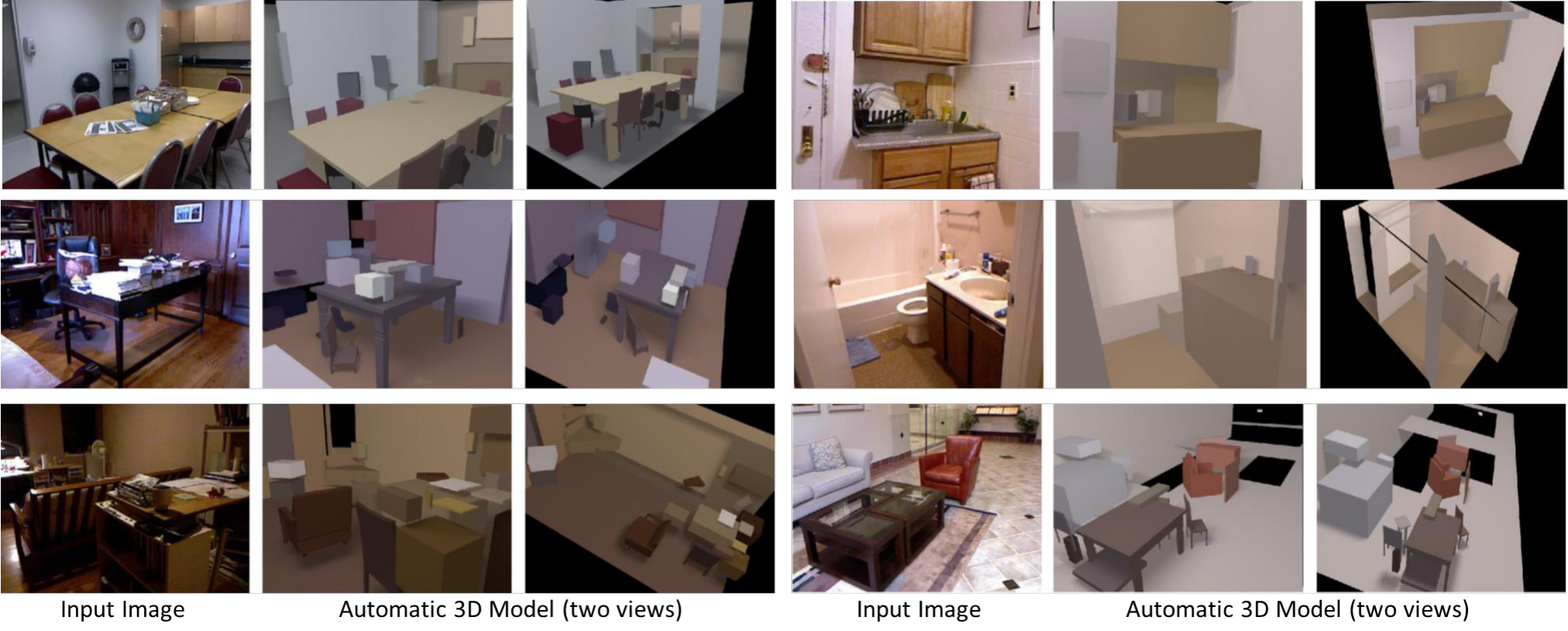}
\vspace{-0.25in}
\caption{
\label{fig:qual_short}
Examples of 3D models estimated automatically from the input RGBD images.  The method is accurate for layout surfaces and large pieces of furniture.  Small and heavily occluded objects remain a major challenge.  Directions for improvement include incorporating object categorization and physical support estimation.
}
\end{figure*}

\section{Experimental Evaluation}

We show representative examples of predicted 3D models in Figures~\ref{fig:qual_long} and~\ref{fig:qual_short}.  We also apply several quantitative measures to evaluate different aspects of our solutions: (1) labels/depth of predicted of layout surfaces; (2) categorization of retrieved regions; (3) voxel accuracy of objects and freespace; (4) instance segmentation induced by rendering.  Among these, the depth accuracy of predicted layout surfaces and voxel accuracy of object occupancy are the most direct evaluations of our 3D model accuracy.  The retrieval categorization serves to evaluate feature suitability and metric learning.  The instance segmentation indicates accuracy of object localization and may be useful for subsequent comparison. To our knowledge, no existing methods produce complete 3D scenes from single RGBD images, so we compare to applicable prior work and sensible baselines on individual measures.

\subsection{Layout Evaluation}
\label{subsec:evallayout}

\begin{figure*}
{\small
\begin{center}
\begin{subfigure}[c]{0.30\textwidth}
\begin{tabular}{|c||c|c|}
\hline
Pixel Err (\%) & NYUParser~\cite{silberman2012eccv} & Ours\\
\hline
Overall & 34.6 & \textbf{10.8} \\
Occluded & 50.0 & \textbf{13.9} \\
Visible & 5.2 &  \textbf{4.8} \\
\hline
\end{tabular}
\subcaption{Layout Error (full dataset, 654 image)\label{fig:layoutlabel}}
\end{subfigure}
\hspace{1.5em}
\begin{subfigure}[c]{0.35\textwidth}
\centering
\begin{tabular}{|c||c|c|}
\hline
Pixel Err (\%) & Zhang et al.~\cite{zhang2013iccv} & Ours\\
\hline
Overall & 10.0 & \textbf{5.4} \\
Occluded & 13.0 & \textbf{7.6} \\
Visible & 5.9 &  \textbf{2.3} \\
\hline
\end{tabular}
\subcaption{Layout Error (on intersection of test subsets in \cite{silberman2012eccv} and \cite{zhang2013iccv}, 47 images)}
\end{subfigure}
\begin{subfigure}[c]{0.28\textwidth}
\vspace{-1.4em}
\begin{tabular}{|c||c|c|}
\hline
DepthErr(m) & Sensor & Ours\\
\hline
Overall & 0.517 & \textbf{0.159} \\
Occluded & 0.739 & \textbf{0.193} \\
Visible & \textbf{0.059} &  0.075 \\
\hline
\end{tabular}
\subcaption{Depth Prediction\label{fig:layoutdepth}}
\end{subfigure}
\end{center}
}
\vspace{-0.2in}
\caption{
\label{fig:layout}
\textbf{Evaluation of Layout}. (a) Pixel labeling error for layout surfaces with 5 categories. (b) Comparison with Zhang et al.~\cite{zhang2013iccv}, note that their layout model are boxes while our ground truth can have non-boxy annotation. (c) Depth error for visible and occluded portions of layouts.  Sensor error is the difference between the input depth image and annotated layout.}
\end{figure*}

In Fig.~\ref{fig:layout}(a,b), we evaluate accuracy of labeling background surfaces into ``left wall'', ``right wall'', ``front wall'', ``ceiling'', and ``floor''.  Ground truth is obtained by rendering the 3D annotation of layout surfaces, and our prediction is obtained by rendering our predicted layout surfaces.  The labels of ``openings'' (e.g., windows that are cut out of walls) are assigned based on the observed depth/normal.  In Fig.~\ref{fig:layout}(a), we compare to the RGBD region classifier of Silberman et al.~\cite{silberman2012eccv} on the full test set.  As expected, we outperform significantly on occluded surfaces (13.9\% vs. 50.0\% error) because ~\cite{silberman2012eccv} does not attempt to find them; we also outperform on visible surfaces (4.8\% vs. 5.2\% error), which is due to the benefit of a structured scene model.  We also compare to Zhang et al.~\cite{zhang2013iccv} who estimate boxy layout from RGBD images on the intersection of their test set with the standard test set (Fig.~\ref{fig:layout}(b)).  These images are easier than average, and our method outperforms substantially, cutting the error nearly in half (5.4\% vs. 10.0\%).

We also evaluate layout depth prediction, the rendered depth of the room without foreground objects (Fig.~\ref{fig:layoutdepth}).  Error is the difference in depth from the ground truth layout annotation.  On visible portions of layout surfaces, the error of our prediction is very close to that of the sensor, with the difference within the sensor noise range.  On occluded surfaces, the sensor is inaccurate (because it measures the foreground depth, rather than that of the background surface), and the average depth error of our method is only 0.193 meters, which is quite good considering that sensor noise is conservatively 0.03*depth.

%

\begin{figure*}
\begin{subfigure}[c]{0.25\textwidth}
\begin{tabular}{|c|c|c|}
\hline
Method & Raw & CCA \\
\hline
3D & 0.169 & 0.226 \\
SIFT & 0.160 & 0.218 \\
CNN & 0.155 & 0.212 \\
\textbf{3D+SIFT} & 0.192 & \textbf{0.265}\\
\hline
\end{tabular}
\subcaption{Metric Learning Evaluation\label{fig:retrievalaccuracy}}
\end{subfigure}
\begin{subfigure}[c]{0.25\textwidth}
\hspace{1em}
\begin{tabular}{|c|c|}
\hline
Feature Set& Accuracy\\
\hline
3D+SIFT & 0.265 \\
\hline
-color &  0.258 \\
-normal & 0.264 \\
-bbox & 0.234 \\
-SIFT & 0.218 \\
\hline
\end{tabular}
\subcaption{Feature Ablation\label{fig:retrievalablation}}
\end{subfigure}
\begin{subfigure}[c]{0.4\textwidth}
\includegraphics[width=\textwidth]{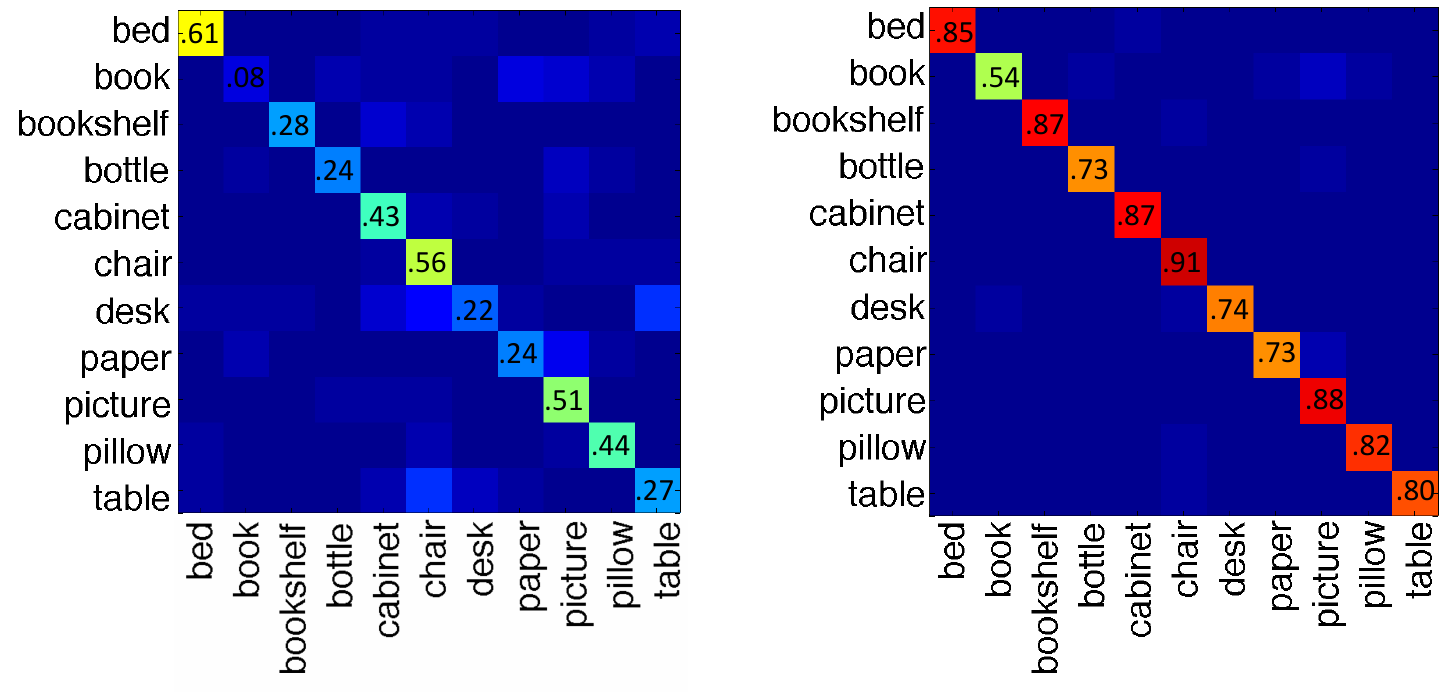}
\subcaption{Confusion Matrix at Top 1 \& Top 10\label{fig:retrievalconfm}}
\end{subfigure}
\vspace{-0.05in}
\caption{
\label{fig:retrieval}
\textbf{Evaluation of Region Retrieval.} (a) Evaluating semantic category prediction with 1-NN under different features and learning approaches (using all categories in NYUv2). (b) Feature ablation test. (c) Top-1 and Top-10 confusion matrix of retrieval. This is a small subset of the confusion matrix with more than $600$ categories, so rows do not sum to 1.
}
\end{figure*}

\subsection{Region Retrieval Evaluation}
\label{subsec:evalretrieval}

To evaluate effectiveness of features and metric learning used for retrieval, we compare categorization accuracy of retrieved regions using different subsets of features.  The features are the {\em 3D features} from~\cite{silberman2012eccv}, including histograms of surface normals, 2D and 3D bounding box dimensions, color histograms and relative depth; {\em SIFT features} from~\cite{silberman2012eccv}; and {\em CNN features} as described by~\cite{girshick2013cnn} using the pre-trained network from~\cite{jia2013caffe}.
In Figure~\ref{fig:retrieval}(a,b), we report the one-nearest-neighbor (1-NN) retrieval accuracy (using all of the hundreds of category labels). The CCA-based metric learning provides a substantial improvement in retrieval accuracy, and the combination of 3D features and RGBD-SIFT provides the best performance.  The set of 3D features, despite their low dimensionality, are the single most useful set of features; within them, bounding box features, such as size and location, are most important.


In Figure~\ref{fig:retrievalconfm}, we show the confusion matrices of some frequent object categories for Top-1 and Top-10 retrieval. The Top-K confusion matrix is computed as such: if there is a correct prediction in Top-K retrieved items, it is considered to be correct; otherwise, it is confused with the highest ranked retrieval.  This shows that at least one of the ten retrieved models for each region is likely to be from the correct semantic category.

\subsection{Occupancy and Freespace Evaluation}
\begin{figure*}[h!t]
\centering
\begin{subfigure}[c]{0.45\textwidth}
\begin{tabular}{|c||c|c|c|}
\hline
Method & Sensor & Ours-Annotated  & Ours-Auto \\
\hline
& & & \\
Precision & \textbf{1.000} & 0.946               & 0.944\\
& & & \\
Recall      & 0.804               &  \textbf{0.932} & 0.926\\
\hline
\end{tabular}
\subcaption{Freespace Evaluation\label{fig:freespace}}
\end{subfigure}
\hspace{2em}
\begin{subfigure}[c]{0.45\textwidth}
\begin{tabular}{|c||c|c|c|}
\hline
Method& Bbox & Ours-Annotated & Ours-Auto\\
\hline
Precision                 & 0.537 & 0.619 & 0.558\\
Recall                      & 0.262 & 0.404 & 0.365\\
Prec.-$\epsilon$     & 0.754 & 0.820 & 0.764\\
Recall-$\epsilon$      & 0.536 &  0.661 & 0.633\\

\hline
\end{tabular}
\subcaption{Occupancy Evaluation\label{fig:occupancy}}
\end{subfigure}
\vspace{-0.05in}
\caption{\label{fig:voxelaccuracy}
\textbf{Evaluation of Voxel Prediction.} (a) Unoccupied voxel precision/recall using our method with ground truth segmentation (Ours-Annotated) or automatic proposals (Ours-Auto), compared to putting voxels around only observed depth points (Sensor); (b) Occupied voxel precision/recall, compared to fitting bounding boxes to ground truth regions.
}
\end{figure*}

We evaluate our scene prediction performance based on voxel prediction. The scope of evaluation is the space surrounded by annotated layout surfaces. Voxels that are out of the viewing scope or behind solid walls are not evaluated. We render objects separately and convert them into a solid voxel map. The occupied space is the union of all the voxels from all objects; free space is the complement of the set of occupied voxels.

Our voxel representation is constructed in a fine grid with $0.03m$ spacing to provide the resolution to encode shape details of 3D model objects we use.  The voxel prediction and recall are presented in Figure~\ref{fig:voxelaccuracy}.  The voxel representation has advantages of being computable from various volumetric representations, viewpoint-invariant, and usable for models constructed from multiple views (as opposed to depth- or pixel-based evaluations).

There is inherent annotation and sensor noise in our data, which is often much greater than $0.03m$.  Objects, when they are small, of nontrivial shape, or simply far away, result in very poor voxel accuracy, even though they agree with the input image. Therefore, we introduce prediction with a tolerance, proportional to the depth of the voxel, for which we use $\epsilon=0.05*depth$, the sensor resolution of Kinect. Specifically, an occupied voxel within $\epsilon$ of a ground truth voxel is considered to be correct (for precision) and to have recalled that ground truth voxel.

We present two simple baselines for comparison. For free space, we evaluate the observed sensor depth. The free space from observed sensor depth predicts 100\% of the visible free space but recalls none of the free space that is occluded. Our estimate recalls 63\% of the occluded freespace with a 5\% drop in precision.  For occupied space, our baseline is generating bounding boxes based on ground truth segmentations with 10\% outlier rejection.  We outperform this baseline, whether using ground truth segmentations or automatic region proposals to generate the model.  Also, note that precision is higher than recall, so it is more common to miss objects than to generate false ones.  Interestingly, the models produced by automatic region proposals (``Ours-Auto'') achieve similar occupancy accuracy as those produced from ground truth proposals (``Ours-Annotated''), showing the effectiveness of generating multiple region and object proposals and selecting among them.

\subsection{Instance Segmentation}

We also evaluate the instance segmentation of our prediction based on 3D rendering of our predicted model (Fig.~\ref{fig:segmentationaccuracy}), following the protocol in RMRC~\cite{rmrc2014}.  Even the ground truth annotations from Guo and Hoiem~\cite{guo2013iccv} (``3D GT'') do not achieve very high performance, because rendered models sometimes do not follow image boundaries well and some small objects are not modeled in annotations.  This provides an upper-bound on our performance.  We compare to the result of Gupta et al.~\cite{gupta2013cvpr}, by applying connected component on their semantic segmentation map.  The result of their more recent paper~\cite{gupta2014eccv} is not available at this time. Since our segmentation is a direct rendering of 3D models, it is more structured and expressive, but tends to be less accurate on boundaries, leading to loss in segmentation accuracy. Segmentation accuracy was not a direct objective of our work, and improving in this area is a possible future direction.

\begin{figure}[ht]
\begin{tabular}{|c||c|c|c|}
\hline
Measure & Gupta et al.~\cite{gupta2013cvpr} & 3D GT & Ours \\
\hline
MeanCovW & 0.533 & 0.583 & 0.505\\
MeanCovU & 0.343 & 0.390 & 0.282\\
\hline
\end{tabular}
\vspace{-0.1in}
\caption{\label{fig:segmentationaccuracy}
\textbf{Instance Segmentation Evaluation} Measures the coverage of ground truth regions and computing the mean over every images, weighted or unweighted by the area of the ground truth region.
}
\vspace{-0.1in}
\end{figure}

\section{Conclusions}
\label{sec:conclusion}

We proposed an approach to predict a complete 3D scene model of individual objects and surfaces from a single RGBD image.  Our results on voxel prediction demonstrate the effectiveness of our proposal-based strategy; our results on layout labeling and depth estimation demonstrate accurate prediction of layout surfaces.  Qualitative results indicate reasonable estimates of occupancy and layout.  Common problems include splitting large objects into small ones, completely missing small objects, not preserving semantics in transferred shapes (not a direct goal), and difficulty with occluded objects, such as chairs.  We see many interesting directions for future work: improving proposals/segmentation (e.g.,~\cite{gupta2014eccv}), incorporating object categories (e.g.,~\cite{song2014eccv,gupta2014eccv}), incorporating support relations (e.g.,~\cite{jia2013cvpr,silberman2014eccv1,silberman2014eccv2}), modeling object context such as chairs tend to be near/under tables, and modeling self-similarity such as that most chairs within one room will look similar to each other.

\vspace{0.05in}
\noindent \textbf{Acknowledgements: } This research is supported in part by ONR MURI grant
N000141010934. 

\newpage
{\small
\bibliographystyle{ieee}
\bibliography{thesisrefs}
}

\end{document}